\documentclass[10pt,twocolumn,letterpaper]{article}

\usepackage{iccv}
\usepackage{times}
\usepackage{epsfig}
\usepackage{graphicx}
\usepackage{amsmath}
\usepackage{amssymb}
\usepackage{bm}
\usepackage{subfigure}
\usepackage{epstopdf}
\usepackage{mathrsfs}
\usepackage[font=small,skip=-5pt]{caption}

\usepackage[breaklinks=true,bookmarks=false]{hyperref}

\iccvfinalcopy 

\def\iccvPaperID{****} 

\ificcvfinal\fi
\begin{document}

\title{Rolling-Shutter-Aware Differential SfM and Image Rectification}

\author{Bingbing Zhuang,~~Loong-Fah Cheong,~~Gim Hee Lee\\
National University of Singapore\\
{\tt\small zhuang.bingbing@u.nus.edu, \{eleclf,gimhee.lee\}@nus.edu.sg}
}

\maketitle
\thispagestyle{empty}

\begin{abstract}
In this paper, we develop a modified differential Structure from Motion (SfM) algorithm that can estimate relative pose from two consecutive frames despite of Rolling Shutter (RS) artifacts.
In particular, we show that under constant velocity assumption, the errors induced by the rolling shutter effect can be easily rectified by a linear scaling operation on each optical flow. We further propose a 9-point algorithm to recover the relative pose
of a rolling shutter camera that undergoes constant acceleration motion. We demonstrate that the dense depth maps recovered from the relative pose of the RS camera  can be used in a RS-aware warping for image rectification to recover high-quality Global Shutter (GS) images.
Experiments on both synthetic and real RS images show that
our RS-aware differential SfM algorithm produces more accurate results on relative pose estimation and 3D reconstruction
from images distorted by RS effect compared
to standard SfM algorithms that assume a GS camera model. We also demonstrate that our RS-aware warping for image rectification method
outperforms state-of-the-art commercial software products, i.e. Adobe After Effects and Apple Imovie, at removing RS artifacts.
\end{abstract}


\section{Introduction}

\begin{figure}
 \setlength{\abovecaptionskip}{-0.35cm}
 \setlength{\belowcaptionskip}{-0.55cm}
\begin{center}
   \includegraphics[width=0.98\linewidth, trim = 20mm 45mm 20mm 10mm, clip]{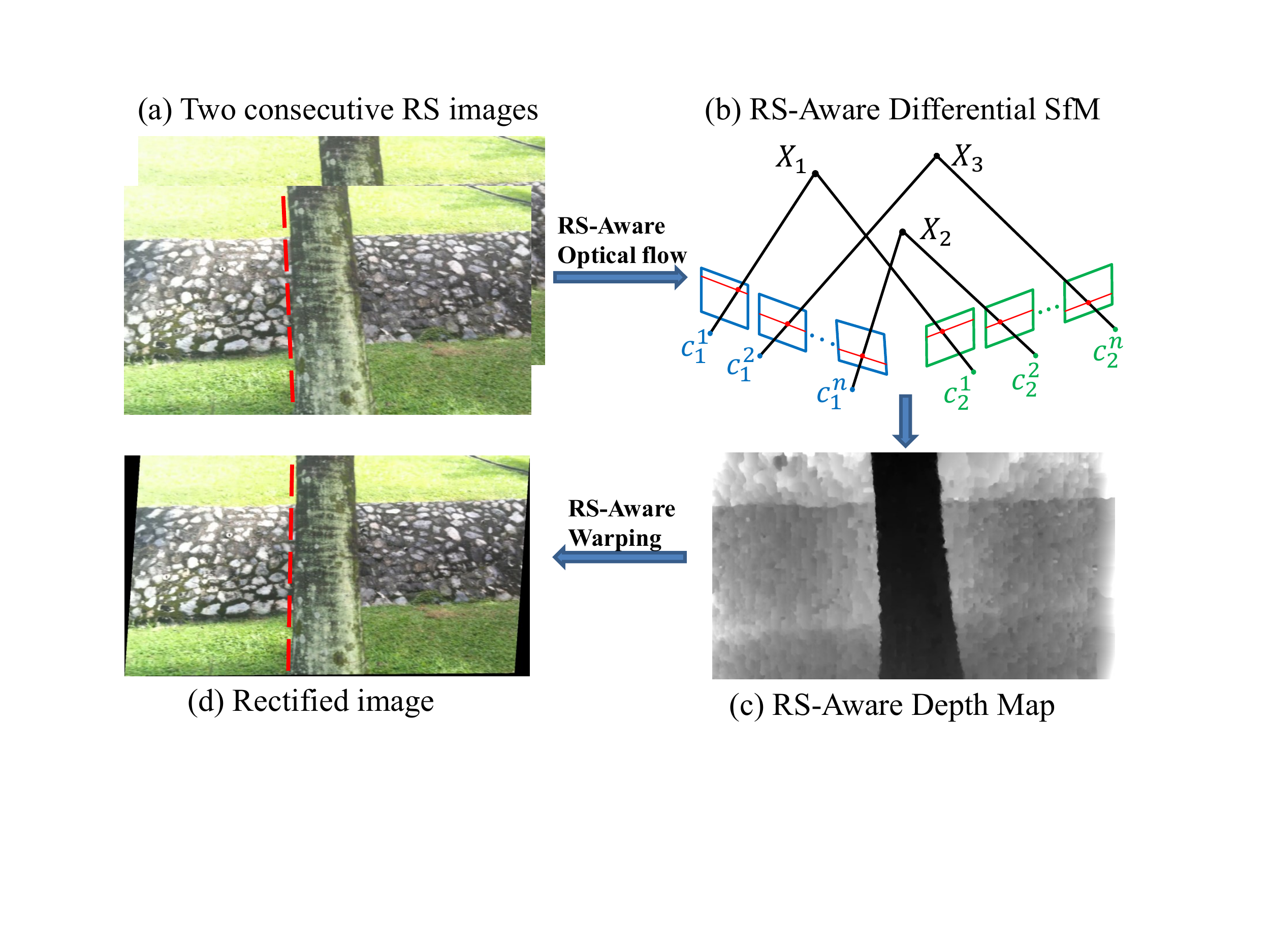}
\end{center}
   \caption{Illustration of our RS-aware differential SfM and image rectification pipeline. See text for more detail.
   }
\label{fig:workflow}
\end{figure}

In comparison with its global shutter (GS) counterpart, rolling shutter (RS) cameras are more widely used
in commercial products due to its low cost. Despite this, the use of RS cameras in computer vision such as
motion/pose estimation is significantly limited compared to the GS cameras.
This is largely due to the fact that most existing computer vision algorithms such as epipolar geometry \cite{Hartley2004} and SfM
\cite{Wu11,Frahm:2010:BuildingRom} make use of the global shutter pinhole camera model which
does not account for the so-called rolling shutter effect caused by camera motion. Unlike a GS camera where the photo-sensor is exposed fully at the same moment,
the photo-sensor of a RS camera is exposed in a scanline-by-scanline fashion due to the exposure/readout modes of the low-cost CMOS sensor.
As a result, the image taken from a moving RS camera is distorted as each scanline possesses a different optical center. An example is shown in Fig.~\ref{fig:workflow}(a), where
the vertical tree trunk in the image captured by a RS camera moving from right to left appears to be slanted.

Due to the price advantage of the RS camera, many researchers began to propose 3D computer vision
algorithms that aim to mitigate the RS effect over the recent years. Although several works have successfully demonstrated
stereo \cite{saurer2013rolling}, sparse and dense 3D reconstruction \cite{Klinger13Streetview,saurersparse} and absolute pose estimation \cite{albl2015r6p,saurer2015IROSrolling,magerand2012global}
using RS images, most of these works completely bypassed the initial relative pose estimation, e.g. by substituting it with GPS/INS readings.
This is because the additional linear and angular velocities of the camera that need to be estimated due
to the RS effect significantly increases the number of unknowns, making the problem intractable.
Thus, despite the efforts from \cite{dai2016rolling} to solve the
RS relative pose estimation problem under discrete motion, the proposed solution is unsuitable for practical use due to the need for high
number of image correspondences that prohibits robust estimation with RANSAC.

In this paper, we aim to correct the RS induced inaccuracies in SfM across two consecutive images under continuous motion by using a differential formulation.
In contrast to the discrete formulation where 12 additional motion parameters from the velocities of the camera need to be solved,  we show that in the differential case, the poses of each scanline can be related to the relative pose of two consecutive images under suitable motion models, thus obviating the need for additional motion parameters.
Specifically, we show that under a constant velocity assumption, the errors induced by the RS effect can be easily rectified by a linear scaling operation on each optical flow, with the scaling factor being dependent on the scanline position of the optical flow vector. To relax the restriction on the motion, we further propose a nonlinear 9-point algorithm for a camera that undergoes constant acceleration motion.  We then apply a RS-aware non-linear refinement step that jointly improves the initial structure and motion estimates by minimizing the geometric errors.
Besides resulting in tractable algorithms, another advantage of adopting the differential SfM formulation lies in the recovery of dense depth map, which we leverage to design a RS-aware warping to remove RS distortion and recover high-quality GS images.
Our algorithm is illustrated in Fig.~\ref{fig:workflow}.
Fig.~\ref{fig:workflow}(a) shows an example of the input RS image pair where a vertical tree trunk appears slanted (red line) from the RS effect, and Fig.~\ref{fig:workflow}(d) shows the vertical tree trunk (red line) restored by our RS rectification. Fig.~\ref{fig:workflow}(b) illustrates the scanline-dependent camera poses for a moving RS camera. $c_i^j$ denotes the optical center of the scanline $j$ in camera $i$. Fig.~\ref{fig:workflow}(c) shows the RS-aware depth map recovered after motion estimation.
Experiments on both synthetic and real RS images validate the utility of our RS-aware differential SfM algorithm.
Moreover, we demonstrate that our RS-aware warping produces rectified images that are superior to popular image/video editing software.

\section{Related works}

Meingast $et~al.$ \cite{Meingast2005} was one of the pioneers to study the geometric model of a rolling shutter camera. Following this work,
many 3D computer vision algorithms have been proposed in the context of RS cameras.
Saurer $et~al.$ \cite{saurersparse} demonstrated large-scale sparse to dense 3D reconstruction using images taken from a
RS camera mounted on a moving car. In another work \cite{saurer2013rolling}, Saurer $et~al.$ showed stereo
results from a pair of RS images. \cite{im2015high} showed high-quality 3D reconstruction from a RS video under small motion.
 Hedborg $et~al.$ \cite{hedborg2012rolling} proposed a bundle adjustment algorithm for RS cameras. In \cite{Klinger13Streetview}, a large-scale bundle adjustment with a generalized camera model is proposed and applied to 3D reconstructions from images collected with a rig of RS cameras. Several works \cite{albl2015r6p,saurer2015IROSrolling,magerand2012global} were introduced to solve the absolute pose estimation problem using RS cameras. All these efforts demonstrated the potential of applying 3D algorithms to RS cameras. However, most of them avoided the initial relative pose estimation problem by taking the information directly from other sensors such as GPS/INS, relying on the global shutter model to initialize the relative pose or assuming known 3D structure.


Recently, Dai $et~al.$ \cite{dai2016rolling} presented the first work to solve the relative pose estimation problem for RS
cameras. They tackled the discrete two-frame relative pose estimation problem by introducing the concept of generalized essential
matrix to account for camera velocities. However, 44 point correspondences are needed to linearly solve for the full motion. This makes the
algorithm intractable when robust estimation via the RANSAC framework is desired for real-world data.
In contrast, we look at the differential motion for two-frame pose estimation where we
show that the RS effect can be compensated in a tractable way.
This model permits a simpler derivation that can be viewed as an extension of conventional optical flow-based differential
pose estimation algorithms \cite{ma2000linear, stewenius2007efficient, zucchelli2002optical} designed
for GS cameras.
Another favorable point for the optical flow-based differential formulation is that unlike the
region-based discrete feature descriptors used in the aforementioned correspondence-based methods, the
brightness constancy assumption used to compute optical flow is not affected by RS distortion as observed in \cite{baker2010removing}.

Several other research attempted to rectify distortions in images caused by the RS effect.
Forss{\'e}n $et~al.$ \cite{ringaby2012efficient} reduced the RS distortion by compensating for 3D camera rotation, which is assumed to be the dominant motion for hand-held cameras.
Some later works \cite{karpenko2011digital,hanning2011stabilizing} further exploited the gyroscope on mobile devices
to improve the measurement of camera rotation.
Grundmann $et~al.$ \cite{grundmann2012calibration} proposed the use of homography mixtures
to remove the RS artifacts. Baker $et~al.$ \cite{baker2010removing} posed the rectification as a temporal super-resolution problem to remove RS wobble. Nonetheless, the distortion is modeled only in the 2D image plane.
To the best of our knowledge, our rectification method is the first that is based on full motion estimation and 3D reconstruction.
This enables us to perform RS-aware warping which returns high-quality rectified images as shown in Sec. \ref{sec:results}.

\section{GS Differential Epipolar Constraint}
\label{sec:GSCamera}

In this section, we give a brief description of the \textit{differential epipolar constraint}
that relates the infinitesimal motion between two \textit{global shutter} camera frames. Since this section does not contain
our contributions, we give only the necessary details
to follow the rest of this paper.
More details of the algorithm can be found in \cite{ma2000linear, zucchelli2002optical}.
Let us denote the linear and angular velocities of the camera by $\bm{v} = [v_x, v_y, v_z]^T$ and $\bm{w} = [w_x, w_y, w_z]^T$.
The velocity field $\bm{u}$ on the image plane induced by $(\bm{v}, \bm{w})$ is given by:
 \vspace{-0.1cm}
\begin{equation}
\bm{u} = \frac{\bm{A}\bm{v}}{Z}+\bm{B}\bm{w},
\label{eq:flowproj}
\end{equation}
 \vspace{-0.6cm}

\noindent where \\ 

 \vspace{-1cm}
\begin{equation}
\setlength{\arraycolsep}{2pt}
\bm{A} = \begin{bmatrix}
   -1 & 0 & x \\
   0 & -1 & y \\
  \end{bmatrix},
\bm{B} = \begin{bmatrix}
   xy & -(1+x^2) & y \\
   (1+y^2) & -xy & -x \\
  \end{bmatrix}.
\end{equation}
 \vspace{-0.3cm}

\noindent $(x,y)$ is the normalized image coordinate and
$Z$ is the corresponding depth of each pixel.
In practice, $\bm{u}$ is approximated by optical flow under brightness constancy assumption.
Given a pair of image position and optical flow vector $(\bm{x},\bm{u})$, Z can be eliminated from Eq.~(\ref{eq:flowproj}) to yield the  differential epipolar constraint
\footnote{Note that our version is slightly different from \cite{ma2000linear} by the sign in the second term due to the difference on how we define the motion.}:
 \vspace{-0.1cm}
\begin{equation}
 \bm{u}^T \hat{\bm{v}}\bm{x} - \bm{x}^T\bm{s}\bm{x} = 0,
 \label{eq:continuousEpipolarGS}
\end{equation}

 \vspace{-0.1cm}
\noindent where $\bm{s}=\frac{1}{2}(\hat{\bm{v}}\hat{\bm{w}}+\hat{\bm{w}}\hat{\bm{v}})$ is a symmetric matrix.
 $\hat{\bm{v}}$ and $\hat{\bm{w}}$ represent the skew-symmetric matrices associated with $\bm{v}$ and $\bm{w}$ respectively.
The space of all the matrices having the same form as $\bm{s}$ is called the \textit{symmetric epipolar space}.
The 9 unknowns from $\bm{v}$ and $\bm{s}$ (3 + 6 respectively) can be solved linearly from at least 8 measurements
$(\bm{x}_j, \bm{u}_j), \forall~j=1,...,8$. The solution returned by the linear algorithm is then projected onto the symmetric epipolar space,
followed by the recovery of ($\bm{v}$,$\bm{w}$) as described in \cite{ma2000linear}.
Note that $\bm{v}$ can only be recovered up to scale.

It is well to remember here that in actual computation, assuming small motion between two frames, all the instantaneous velocity terms will be approximated by displacement over time. Removing the common factor of time, the optical flow vector $\bm{u}$ now indicates the displacement of pixel over two consecutive frames, and the camera velocities $(\bm{v}, \bm{w})$ indicate the relative pose of the two camera positions.
 The requisite mapping between the relative orientation $\bm{R}$ and the angular velocity $\bm{w}$ is given by the well-known mapping $\bm{R}\!=\!\exp(\bm{w})\simeq \bm{I}+\hat{\bm{w}} $.
Henceforth, we will call $(\bm{v}, \bm{w})$ relative motion/pose for the rest of this paper. We utilize Deepflow \cite{weinzaepfel2013deepflow} to compute the optical flow for all our experiments
due to its robust performance in practice.


\section{RS Differential Epiplor Constraint}
\subsection{Constant Velocity Motion}
\begin{figure}[h]
 \setlength{\belowcaptionskip}{-0.45cm}
 \setlength{\abovecaptionskip}{-0.35cm}
\begin{center}
   \includegraphics[width=1\linewidth, trim = 28mm 77mm 30mm 66mm, clip]{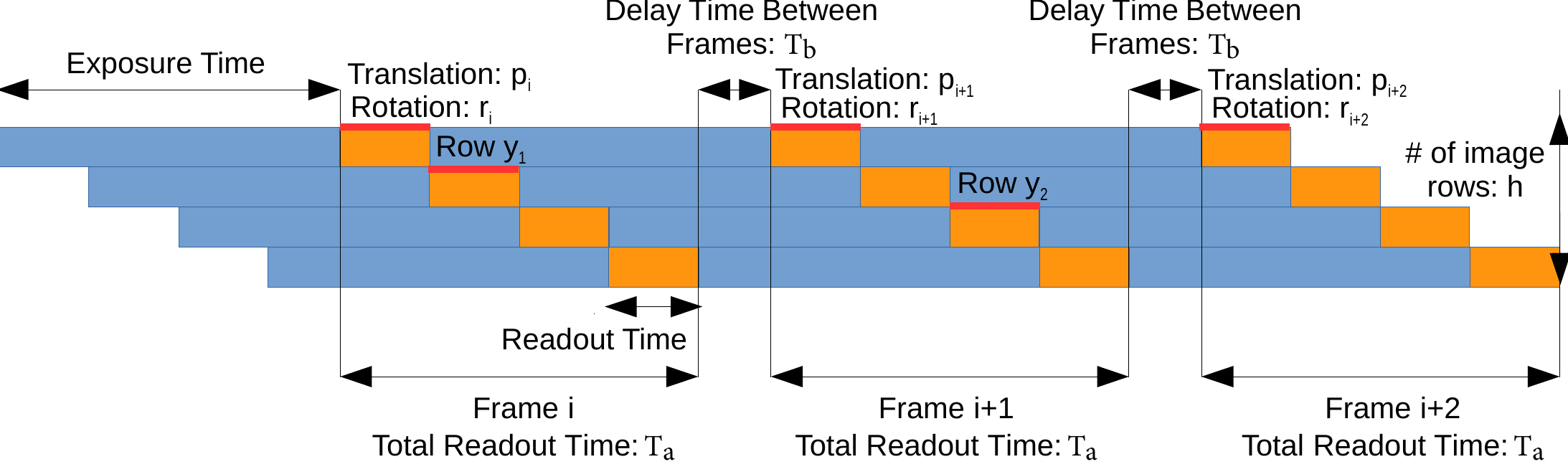}
\end{center}
   \caption{Illustration of exposure, readout and delay times in a rolling shutter camera.}
\label{fig:rotationInterpolation}
\end{figure}

The differential epipolar constraint shown in the previous
section works only on images taken with a GS camera and would fail if the images
were taken with a RS camera. The main difference between the RS and GS
camera is that we can no longer regard each image as having one single camera pose. Instead we have to introduce
a new camera pose for each scanline on a RS image as shown in Fig.~\ref{fig:workflow}(b) due to the readout and delay times as
illustrated in Fig.~\ref{fig:rotationInterpolation}.
 \vspace{-0.1cm}

Consider three consecutive image frames $i, i+1$ and $i+2$. Let $\{\bm{p}_i,~\bm{p}_{i+1},~\bm{p}_{i+2}\}$ and
$\{\bm{r}_i,~\bm{r}_{i+1},~\bm{r}_{i+2}\} \in so(3)$
represent the translation and rotation of the first scanlines on the respective images as shown in Fig.~\ref{fig:rotationInterpolation}.
Frame $i$ is set as the reference frame, i.e. $(\bm{p}_i,\bm{r}_i) = (\bm{0},\bm{0})$.
Now consider an optical flow which maps an image point from $(x_1,y_1)$ in frame $i$ to $(x_2,y_2)$ in frame $i+1$.
Assuming constant instantaneous velocity of the camera across these three frames under small motion, we can compute the translation and rotation $(\bm{p}_1, \bm{r}_1)$
of scanline
$y_1$ on frame $i$ as a linear interpolation between the poses of the first scanlines from frames $i$ and $i+1$:
 \vspace{-0.3cm}
\begin{subequations}
\begin{equation}
 \bm{p}_1 = \bm{p}_i + \frac{\gamma y_1}{h}(\bm{p}_{i+1} - \bm{p}_{i}),
\end{equation}
 \vspace{-0.25cm}
\begin{equation}
 \bm{r}_1 = \bm{r}_i + \frac{\gamma y_1}{h}(\bm{r}_{i+1} - \bm{r}_{i}).
\end{equation}
\label{eq:poseScanline_y1}
\end{subequations}

 \vspace{-0.25cm}
\noindent $h$ is the total number of
scanlines in the image.  $\gamma = \frac{T_a}{T_a+T_b}$ is the readout time ratio which can be obtained a priori
from calibration \cite{Meingast2005}.
Similar interpolation for the pose $(\bm{p}_2, \bm{r}_2)$  of the scanline $y_2$ on frame $i+1$ can be done
between the first scanlines from frames $i+1$ and $i+2$:
 \vspace{-0.05cm}
\begin{subequations}
\begin{equation}
 \bm{p}_2 = \bm{p}_{i+1} + \frac{\gamma y_2}{h}(\bm{p}_{i+2} - \bm{p}_{i+1}),
\end{equation}
 \vspace{-0.3cm}
\begin{equation}
 \bm{r}_2 = \bm{r}_{i+1} + \frac{\gamma y_2}{h}(\bm{r}_{i+2} - \bm{r}_{i+1}).
\end{equation}
\label{eq:poseScanline_y2}
\end{subequations}

 \vspace{-0.25cm}
\noindent Now we can obtain the relative motion $(\bm{p}_{21}, \bm{r}_{21})$ between the two scanlines $y_2$ and $y_1$ by taking the difference
of Eq.~(\ref{eq:poseScanline_y2}) and (\ref{eq:poseScanline_y1}), and setting $(\bm{p}_{i+2}-\bm{p}_{i+1})=(\bm{p}_{i+1}-\bm{p}_{i})$ and
$(\bm{r}_{i+2}-\bm{r}_{i+1})=(\bm{r}_{i+1}-\bm{r}_{i})$ due to the constant velocity assumption:
 \vspace{-0.25cm}
\begin{subequations}
\begin{equation}
\begin{split}
  \bm{p}_{21} &= \underbrace{\left( 1 + \frac{\gamma}{h}(y_2 - y_1)\right)}_{\alpha}(\bm{p}_{i+1} - \bm{p}_{i})  \\
\Rightarrow  \bm{p}_{21} &= \alpha (\bm{p}_{i+1} - \bm{p}_{i}),
\end{split}
\end{equation}
\begin{equation}
\begin{split}
  \bm{r}_{21} &= \underbrace{\left( 1 + \frac{\gamma}{h}(y_2 - y_1)\right)}_{\alpha}(\bm{r}_{i+1} - \bm{r}_{i}) \\
\Rightarrow  \bm{r}_{21} &=  \alpha (\bm{r}_{i+1} - \bm{r}_{i}).
\end{split}
\end{equation}
\label{eq:scanlinePose}
\end{subequations}

 \vspace{-0.15cm}
\noindent $\alpha$ is the dimensionless scaling factor of the relative pose made up of $\gamma$, $h$, $y_1$ and $y_2$. It was mentioned in the previous section that under small motion, $({\bm{v}}, {\bm{w}})$ in the differential epipolar constraint can be regarded as the relative pose of the camera in practice.
We can thus substitute $({\bm{v}}, {\bm{w}})$ from Eq.~(\ref{eq:continuousEpipolarGS}) with the relative pose $(\bm{p}_{21}, \bm{r}_{21})$ from Eq.~(\ref{eq:scanlinePose}).
Consequently, we get the rolling shutter differential epipolar constraint
 \vspace{-0.3cm}
\begin{equation}
 \frac{\bm{u}^T}{\alpha} \hat{\bm{v}}_g \bm{x} - \bm{x}^T \bm{s}_g \bm{x} = 0,
 \label{eq:continuousEpipolarRS_Nodt2}
\end{equation}

 \vspace{-0.1cm}
\noindent where $ \bm{s}_g = \frac{1}{2}(\hat{\bm{v}}_g\hat{\bm{w}}_g+\hat{\bm{w}}_g\hat{\bm{v}}_g)$,
$\bm{v}_g = (\bm{p}_{i+1} - \bm{p}_{i})$ and $\bm{w}_g = (\bm{r}_{i+1} - \bm{r}_{i})$.  $\bm{v}_g$ and $\bm{w}_g$ describe the relative pose between the first scanlines
 of two rolling shutter frames, and can be taken to be the same as $\bm{v}$ and $\bm{w}$ from the global shutter case.
It can be seen from Eq.~(\ref{eq:continuousEpipolarRS_Nodt2}) that our
differential epipolar constraint for rolling shutter cameras differs from the differential epipolar constraint for global
shutter cameras (Eq.~(\ref{eq:continuousEpipolarGS})) by just the scaling factor $\alpha$ on the optical flow vector $\bm{u}$.
Here, we can make the interpretation that the rolling shutter optical flow vector $\bm{u}$ when scaled by $\alpha$ is equivalent to the global shutter optical flow vector. Collecting all optical flow vectors, and rectifying each of them with its own $\alpha$ (dependent on the scanlines involved in the optical flow), we can now solve for the RS relative motion using conventional linear 8-point algorithm \cite{ma2000linear}.

\subsection{Constant Acceleration Motion}
Despite the simplicity of compensating for the RS effect by scaling the measured optical flow vector,
the constant velocity assumption can be too restrictive for real image sequences captured by a moving RS camera.
To enhance the generality of our model, we relax the constant velocity assumption to the more realistic constant acceleration motion.
More specifically, we assume constant direction of translational and rotational velocity, but allow its magnitude to either increase or decrease gradually.
Experimental results on real data show that this relaxation on motion assumption improves the performance significantly.

The constant acceleration model slightly complicates the interpolation for the pose of each scanline, compared to the constant velocity model.
We show only the derivations for the translation of the scanlines since similar derivations apply to the rotation.
Suppose the initial translational velocity of the camera at $\bm{p}_i$ is $\bm{\mathcal{V}}$ and it maintains a constant acceleration $\bm{a}$
such that at time $t$ the velocity increases or decreases to $\bm{\mathcal{V}} + \bm{a}t $, and the translation $\bm{p}(t)$ is
 \vspace{-0.15cm}
\begin{equation}
\bm{p}(t) = \bm{p}_i+\int_{0}^{t}(\bm{\mathcal{V}}+\bm{a}t^\prime)dt^\prime = \bm{p}_i +\bm{\mathcal{V}}t+\frac{1}{2}\bm{a}t^2.
\label{eq:accumulated}
\end{equation}

 \vspace{-0.15cm}
\noindent Let us re-parameterize $\bm{\mathcal{V}}$ and $\bm{a}$ as $\bm{\mathcal{V}} = \frac{\Delta \bm{p}}{\Delta t}$ and $\bm{a} = k\frac{\bm{\mathcal{V}}}{\Delta t}$,
where  $\Delta \bm{p}$ is an auxiliary variable introduced to represent a translation, $\Delta t $ is the time period between two first scanlines, and $k$ is a scalar factor
that needs to be estimated.
Putting $\bm{\mathcal{V}}$, $\bm{a}$ back into Eq.~(\ref{eq:accumulated}) and let $t = \Delta t$, we get the translation for the first scanline of frame $i+1$ as
 \vspace{-0.2cm}
\begin{equation}
 \bm{p}_{i+1} =   \bm{p}_i + (1+\frac{1}{2}k)\Delta\bm{p}.
 \label{eq:translation}
\end{equation}

 \vspace{-0.1cm}
\noindent Denoting the time stamp of scanline $y_1$ (or $y_2$) on image $i$
 (or $i+1$) by $t_{y_1}$ (or $t_{y_{2}}$), we have

\begin{equation}
   t_{y_1} = \frac{\gamma y_1}{h}\Delta t,~~~~~t_{y_{2}} = (1+\frac{\gamma y_2}{h})\Delta t.
  \label{eq:timestamp}
\end{equation}


\noindent Substituting the two time instances in Eq.~(\ref{eq:timestamp}) into Eq.~(\ref{eq:accumulated}) and eliminating $\Delta \bm{p}$ by  Eq.~(\ref{eq:translation})  gives rise to the translations of scanline $y_1$ and $y_2$:

\begin{subequations}
\vspace{-0.4cm}
\begin{align}
 \bm{p}_1 &= \bm{p}_i +  \frac{\gamma y_1}{h}\Delta \bm{p} + \frac{1}{2}k(\frac{\gamma y_1}{h})^2\Delta \bm{p} \label{eq:acceposition_a}\\
          &= \bm{p}_i + \underbrace{(\frac{\gamma y_1}{h} + \frac{1}{2}k(\frac{\gamma y_1}{h})^2)(\frac{2}{2+k})}_{\beta_1(k)}(\bm{p}_{i+1} -  \bm{p}_i),  \notag
\end{align}

 \vspace{-0.43cm}
\begin{align}
 \bm{p}_2 &= \bm{p}_i + (1+\frac{\gamma y_2}{h})\Delta \bm{p} + \frac{1}{2}k(1+\frac{\gamma y_2}{h})^2\Delta \bm{p}\\
          &= \bm{p}_i + \underbrace{(1+\frac{\gamma y_2}{h} + \frac{1}{2}k(1+\frac{\gamma y_2}{h})^2)(\frac{2}{2+k})}_{\beta_2(k)}(\bm{p}_{i+1} -  \bm{p}_i). \notag
\end{align}
\end{subequations}

 \vspace{-0.15cm}
\noindent Similar to Eq.~(\ref{eq:scanlinePose}), we get the relative translation and rotation between scanline $y_2$ and $y_1$ as follows:
 \vspace{-0.1cm}
\begin{equation}
\bm{p}_{21} = \beta(k)(\bm{p}_{i+1}-\bm{p}_i),~~~\bm{r}_{21} = \beta(k)(\bm{r}_{i+1}-\bm{r}_i),
\label{eq:accerelative}
\end{equation}

\noindent where $\beta(k) = \beta_2(k)-\beta_1(k)$. Making use of the small motion assumption, we plug $(\bm{p}_{21}$, $\bm{r}_{21})$ into Eq.~(\ref{eq:continuousEpipolarGS}) and
the RS differential epipolar constraint can now be written as
 \vspace{-0.1cm}
\begin{equation}
 \bm{u}^T\hat{\bm{v}}_g \bm{x} - \beta(k)\bm{x}^T \bm{s}_g \bm{x} = 0.
 \label{eq:continuousEpipolarRS_acce}
\end{equation}

 \vspace{-0.0cm}
\noindent It is easy to verify that Eq.~(\ref{eq:continuousEpipolarRS_acce}) reduces to Eq.~(\ref{eq:continuousEpipolarRS_Nodt2}) when the acceleration vanishes , i.e. $k=0$ (constant velocity).

In comparison to the constant velocity model, we have one additional unknown motion parameter $k$ to be estimated, making Eq.~(\ref{eq:continuousEpipolarRS_acce}) a polynomial equation.
In what follows, we show that Eq.~(\ref{eq:continuousEpipolarRS_acce}) can be solved by a 9-point algorithm
with the hidden variable resultant method \cite{cox2007ideals}. Rewriting $\bm{v}_g$ as $[v_x,v_y,v_z]^T$ and the symmetrical matrix $\bm{s}_g$ as

 \vspace{-0.10cm}
$$
 \bm{s}_g = \begin{bmatrix}
   s_1 & s_2 & s_3 \\
   s_2 & s_4 & s_5 \\
   s_3 & s_5 & s_6
  \end{bmatrix},
$$

 \vspace{-0.0cm}
\noindent Eq.~(\ref{eq:continuousEpipolarRS_acce}) can be rearranged to
 \vspace{-0.15cm}
\begin{equation}
z(k)\bm{e} = 0,
\label{eq:poly}
\end{equation}

 \vspace{-0.15cm}
\noindent where $z(k)$ is a $1\times9$ vector made up of the known variables $\gamma,h,\bm{x}$ and $\bm{u}$, and the unknown variable $k$.
$\bm{e}$ is a $9\times1$ unknown vector as follows:
 \vspace{-0.15cm}
 \begin{equation}
\begin{aligned}
 \bold{e} = [ & v_x,~v_y,~v_z,~s_1,~s_2,~s_3,~s_4,~s_5,~s_6]^T.
\end{aligned}
\label{eq:e}
\end{equation}

 \vspace{-0.1cm}
 \noindent We need 9 image position and optical flow vectors $(\bm{x},\bm{u})$ to determine the 10 unknown variables $k$ and $\bm{e}$ up to a scale.
 Each point yields one constraint in the form of  Eq.~(\ref{eq:poly}). Collecting these constraints from all the points, we get a polynomial system:
 \vspace{-0.1cm}
 \begin{equation}
 \bm{\mathcal{Z}}(k)\bm{e} = 0,
 \label{eq:ploysystem}
 \end{equation}

 \vspace{-0.1cm}
  \noindent where $\bm{\mathcal{Z}}(k) = [z_1(k)^T,z_2(k)^T,...,z_9(k)^T]^T$ is a $9\times9$ matrix.
 For Eq.~(\ref{eq:ploysystem}) to have a non-trivial solution, $\bm{\mathcal{Z}}(k)$ must be rank-deficient which implies a vanishing determinant:
\vspace{-0.15cm}
 \begin{equation}
 det(\bm{\mathcal{Z}}(k)) = 0.
 \label{eq:det}
 \end{equation}

\vspace{-0.2cm}
 \noindent Eq.~(\ref{eq:det}) yields a 6-degree univariate polynomial in terms of the unknown $k$ which can be solved by the technique of Companion matrix \cite{cox2007ideals} or Sturm bracketing \cite{nister2004efficient}.
 Next, the Singular Value Decomposition (SVD) is applied to $\bm{\mathcal{Z}}(k)$, and the singular vector associated with the least singular value is
 taken to be $\bm{e}$. Following \cite{ma2000linear}, we extract $(\bm{v}_g,\bm{w}_g)$ from $\bm{e}$ by a projection onto the symmetric epipolar space. The minimal solver takes less than 0.02s using our unoptimized MATLAB code.

\subsection{RS-Aware Non-Linear Refinement}
\label{sec:RSBA}
It is clear that the above algorithm minimizes the algebraic errors and thus yields a biased solution. To obtain more accurate solution, this should be followed by one more step of non-linear refinement that minimizes the geometric errors.
In the same spirit of re-projection error in the discrete case and combining Eq.~(\ref{eq:flowproj}) and (\ref{eq:accerelative}),
 we write the differential re-projection error and non-linear refinement as
\vspace{-0.2cm}
\begin{equation}
\underset{k, \bm{v}_g, \bm{w}_g,\bm{Z}}{\operatorname{argmin}} \sum_{i\in O}^{N}||\bm{u}_i -\beta^i(k)(\frac{\bm{A}_i\bm{v}_g}{Z_i}+\bm{B}_i\bm{w}_g)||_2^2,
\label{eq:nonlinear}
\end{equation}

\vspace{-0.1cm}
\noindent which minimizes the errors between the measured and predicted
optical flows for all points in the pixel set $O$ over the estimated parameters $k$, $\bm{v}_g$, $\bm{w}_g$, $\bm{Z}= \{Z_1, Z_2, ... Z_N\}$. $\bm{Z}$
is the depths associated with all the image points in $O$.
$N$ is the total number of points.
Note that in the case of constant velocity model, $k$ is kept fixed as zero in this step. Also note that (\ref{eq:nonlinear}) reduces to the traditional non-linear refinement for GS model \cite{bruss1983passive,zucchelli2002maximum,hu2009linear} when the readout time ratio $\gamma$ is set as 0.
RANSAC is used to obtain a robust initial estimate. For each RANSAC iteration, we apply our minimal solver to obtain $k$, $\bm{v}_g$, $\bm{w}_g$ and then compute the optimal depth for each pixel by  minimizing (\ref{eq:nonlinear}) over $\bm{Z}$; the inlier set is identified by checking the resultant differential re-projection error on each pixel. The threshold is set as 0.001 on the normalized image plane for all experiments.
We then minimize (\ref{eq:nonlinear}) for all points in the largest inlier set from RANSAC to improve the initial estimates by block coordinate descent
over $k$, $\bm{v}_g$, $\bm{w}_g$ and $\bm{Z}$, whereby each subproblem block admits a closed-form solution. Finally, $\bm{Z}$ is recovered for all pixels which gives the dense depth map.

\section{RS-Aware Warping For Image Rectification}
\label{sec:undist}
Having obtained the camera pose for each scanline and the depth map of the first RS image frame,
a natural extension is to take advantage of these information to rectify the image distortion caused by the RS effect.
From Eq.~(\ref{eq:acceposition_a})
we know that the relative poses $(\bm{p}_{1i},\bm{r}_{1i})$ between the first and other scanlines in the same image are as follow:
 \vspace{-0.2cm}

 \begin{subequations}
\begin{equation}
\bm{p}_{1i} =  \bm{p_1}-\bm{p_i} = \beta_1(k)\bm{v}_g,
\end{equation}
 \vspace{-0.45cm}
\begin{equation}
\bm{r}_{1i} = \bm{r_1}-\bm{r_i} = \beta_1(k)\bm{w}_g.
\end{equation}
\end{subequations}

 \vspace{-0.05cm}
\noindent Combining the pose of each scanline with the depth map, warping can be done by back-projecting each pixel on each scanline into the 3D space, which gives the point cloud, followed by a projection onto
the image plane that corresponds to the first scanline.
Alternatively, the warping displacement can be computed from Eq.~(\ref{eq:flowproj}) by small motion approximation as $\bm{u}_w = \beta_1(k)(\frac{\bm{A}\bm{v}_g}{Z}+\bm{Bw}_g)$.

Since the camera positions of each scanline within the same image are fairly close to that of the first scanline,
the displacement caused by the warping is small compared to the optical flow between two consecutive frames.
Thus warping-induced gaps are negligible and we do not need to use any pixel from the next frame (i.e. image $i+1$) for the rectification.
This in turn means that the warping introduces no ghosting artifacts caused by misalignment, allowing the resulting image to retain the sharpness of the original image while removing the geometric RS distortion, as shown by the experimental results in Sec.~\ref{sec:results}.

\section{Experiments}
\label{sec:results}

In this section, we show the experimental results of our proposed algorithm on both synthetic and real image data.

\subsection{Synthetic Data}
\begin{figure}
 \setlength{\abovecaptionskip}{-0.25cm}
 \setlength{\belowcaptionskip}{-0.45cm}
\begin{center}
   \includegraphics[width=1\linewidth, trim = 8mm 125mm 39mm 0mm, clip]{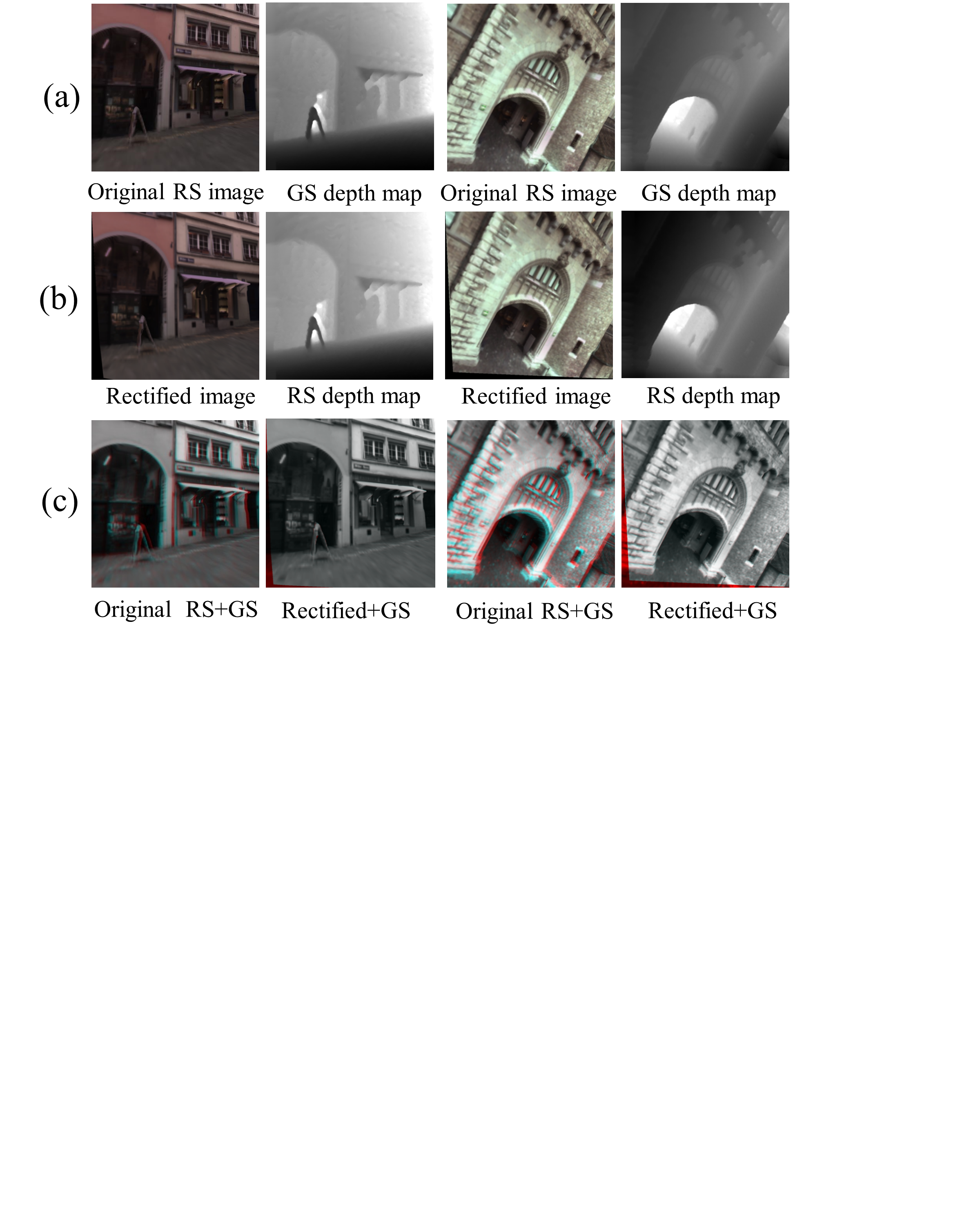}
\end{center}
   \caption{An example of the experimental results on the \textit{Old Town} and \textit{Castle} data. (a)-(b): The original RS images, estimated depth maps by GS \& RS, and rectified images. (c) Overlaying the original RS and the rectified images on the ground truth GS images.}
\label{fig:ptcloudsyn1}
\end{figure}

We generate synthetic but realistic RS images by making use of two textured 3D mesh---the \textit{Old Town} and \textit{Castle} provided by \cite{saurer2013rolling} for our experiments.
To simulate the RS effect, we first use the 3D Renderer software \cite{hassner2013viewing} to render the GS images according to the pose of scanlines. From these GS images,
we extract the pixel values for each scanline to generate the RS images.
As such, we can fully control all the ground truth camera and motion parameters including the readout time ratio $\gamma_{\scriptscriptstyle G}$,
 camera relative translation $\bm{v}_{\scriptscriptstyle G}$ and rotation
$\bm{R}_{\scriptscriptstyle G} = \exp(\bm{w}_{\scriptscriptstyle G}) $,
and acceleration parameter $k_{\scriptscriptstyle G}$ in the case of constant acceleration motion.
The image size is set as $900\times900$ with a 810 pixels focal length.
Examples of the rendered RS images from both datasets are shown in the first row of Fig.~\ref{fig:ptcloudsyn1}.
For the  relative motion estimate $({\bm{v}_g},{\bm{w}_g})$, we measure the translational error as $\cos ^{-1}(\bm{v}_g^T{\bm{v}_{\scriptscriptstyle G}}/( \left \|\bm{v}_g\right \| \left \|{\bm{v}_{\scriptscriptstyle G}}\right \|))$
and the rotational error as the norm of the Euler angles from $\bm{R}_g{\bm{R}_{\scriptscriptstyle G}^T}$, where $\bm{R}_g = \exp({\bm{w}_g})$.
Since the translation is ambiguous in its magnitude,
the amount of translation is always represented as the ratio between the absolute translation magnitude and average scene depth in the rest of this paper.
We term this ratio as normalized translation.


\begin{figure}
 \setlength{\abovecaptionskip}{-0.4cm}
 \setlength{\belowcaptionskip}{-0.6cm}
\begin{center}
   \includegraphics[width=0.95\linewidth,trim = 8mm 30mm 39mm 130mm, clip]{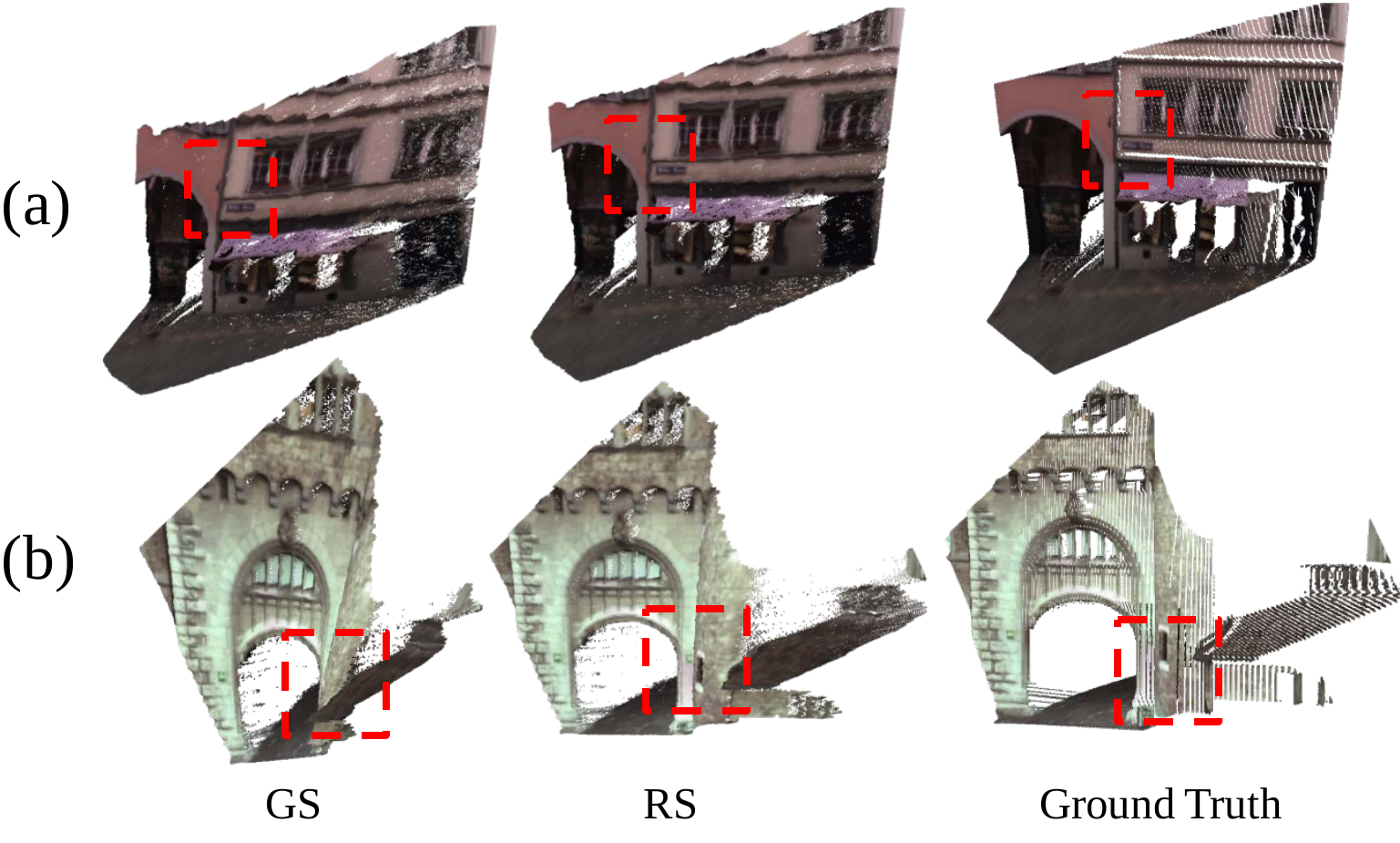}
\end{center}
   \caption{Visualization of the reconstructed 3D point clouds.}
\label{fig:ptcloudsyn2}
\end{figure}

\textbf{Quantitative Evaluation:}  We compare the accuracy of our RS-aware motion estimation to the conventional GS-based model.
We avoid forward motion which is well-known to be challenging for SfM even for traditional GS cameras. WLOG, all the motions that we synthesize have equal vertical and horizontal translation components, and equal yaw, pitch and roll rotation components.
To fully understand the behavior of our proposed algorithm, we investigate the performance under various settings. To get statistically meaningful result, all the errors are obtained from an average
of 100 trials, each with 300 iterations of RANSAC. Both the results from the minimal solver and the non-linear refinement are reported to study their respective contribution to the performance.

\begin{figure*}
\setlength{\abovecaptionskip}{-0.25cm}
 \setlength{\belowcaptionskip}{-0.55cm}
\begin{center}
   \includegraphics[width=1\linewidth, trim = 68mm 2mm 73mm 3mm, clip]{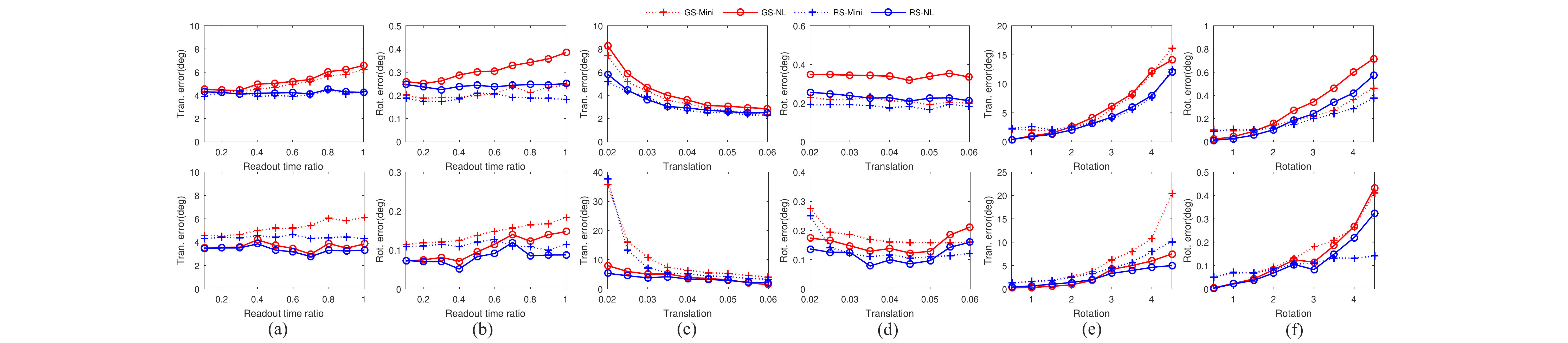}
\end{center}
\vspace{-0.1cm}
   \caption{Quantitative evaluation for constant velocity motion on \textit{Old Town} (first row) and \textit{Castle} (second row). GS-Mini/RS-Mini and GS-NL/RS-NL stand for the results from the minimal solver and non-linear
   refinement respectively using GS/RS model.}
\label{fig:comrsgs_constant}
\end{figure*}

We plot the translational and rotational error 
under the constant velocity motion in Fig.~\ref{fig:comrsgs_constant}.
We first investigate how the value of the readout time ratio $\gamma$ would affect the
performance in Fig.~\ref{fig:comrsgs_constant}(a)-(b) by increasing $\gamma$ from $0.1$ to $1$, while the normalized translation and magnitude of $\bm{w}$ are fixed at 0.025 and $3^\circ$ respectively.
We can see that the accuracy of the RS model (both minimal solver and non-linear refinement) is insensitive to the variation of $\gamma$, while the GS model tends to give higher errors with increasing $\gamma$.
This result is expected because a larger readout time ratio leads to larger RS distortion in the image. Next,
we fixed the value of $\gamma$ to 0.8 for the following two settings:
(1) We fix the magnitude of $\bm{w}$ to $3^\circ$ and increase the normalized translation from 0.02 to 0.06 as shown in Fig.~\ref{fig:comrsgs_constant}(c)-(d). (2)
The normalized translation is fixed as 0.025 and the magnitude of $\bm{w}$ is increased from $0.5^\circ$ to $4.5^\circ$ as shown in Fig.~\ref{fig:comrsgs_constant}(e)-(f).
Overall, the accuracies of the RS and GS model have a common trend determined by the type of motion.
However, the RS model has higher accuracies in general, especially in the challenging cases where the rotation is relatively large compared to translation.
This implies that our RS-aware algorithm has compensated for the RS effect in pose estimation.
We note that in some cases, especially in \textit{Old Town}, the non-linear refinement gives marginal improvement or even increases the error slightly. We reckon this is because the RANSAC criterion we used
is exactly the individual term that forms the non-linear cost function, and it can happen that the initial solution is already close to a local minimum,
hence the effect of non-linear refinement can become dubious given that Eq.(\ref{eq:flowproj}) is only an approximation for small discrete motion in practice, as mentioned in Sec.\ref{sec:GSCamera}.

\begin{figure}
\setlength{\abovecaptionskip}{-0.2cm}
 \setlength{\belowcaptionskip}{-0.7cm}
\begin{center}
   \includegraphics[width=0.86\linewidth, trim = 5mm 25mm 23mm 10mm, clip]{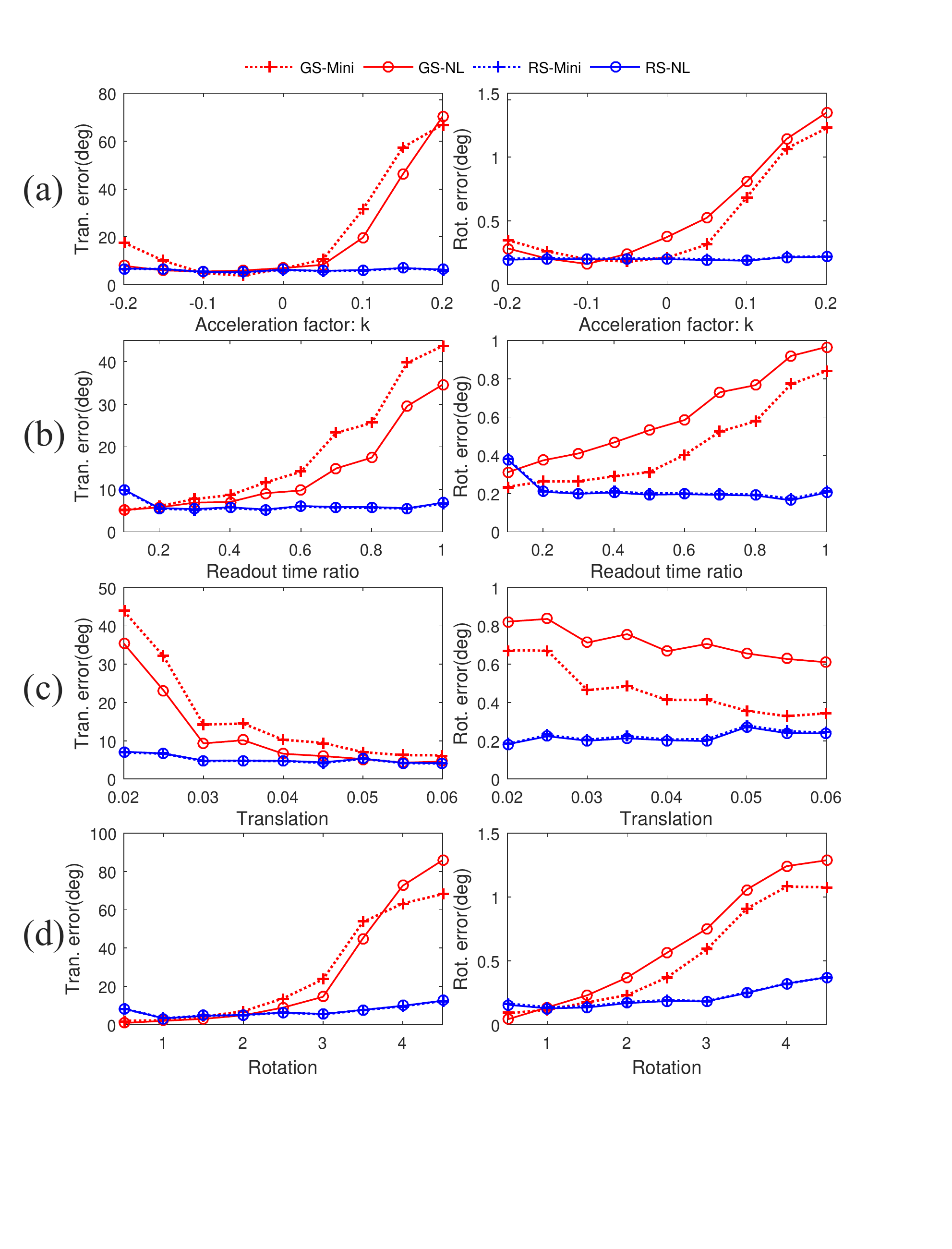}
\end{center}
\vspace*{-5mm}
   \caption{Quantitative evaluation for constant acceleration motion on \textit{Old Town}. The legend is the same as in Fig.\ref{fig:comrsgs_constant}.}
\label{fig:comrsgs_acce}
\end{figure}

Similarly, we conduct quantitative evaluations under the constant acceleration motion. To save space, only the results from \textit{Old Town} are reported here. See \textit{supplementary material} for 
the similar results from \textit{Castle}.
First, we investigate how the variation of acceleration by increasing $k$ from $-0.2$ to $0.2$ would influence the performance of both the GS and RS model in Fig.~\ref{fig:comrsgs_acce}(a).
We can see that the accuracy of the GS model degrades dramatically under large acceleration, while the RS model maintains almost consistent accuracies regardless of the amount of acceleration.
For Fig.~\ref{fig:comrsgs_acce}(b)-(d), we fix $k$ to 0.1 and set other motion or camera parameters to be the same as that for the constant velocity motion.
As can be observed, the RS model in general yields higher accuracies than the GS model, especially for the translation.
For example, the GS model gives significantly larger error ($>50^\circ$) on translation under strong rotation as shown in Fig.~\ref{fig:comrsgs_acce}(d).
We observe that the non-linear refinement tends to improve the translation estimate but degrade the rotation estimate for the GS model. For the RS model, the impact is marginal.
 We observe larger improvement when the RANSAC threshold is increased, but this leads to a drop of overall accuracy.
See our \textit{supplementary material} for more analyses on the quantitative results.


For qualitative results, two examples under constant acceleration motion are shown in Fig.\ref{fig:ptcloudsyn1}\&\ref{fig:ptcloudsyn2}.
Fig.~\ref{fig:ptcloudsyn1}(a)\&(b) show the original synthetic RS images, the estimated depth maps using the GS model, and the estimated depth maps and rectified images using our RS model.
In Fig.~\ref{fig:ptcloudsyn1}(c), we compare the original RS images and rectified images to the ground truth GS images, which are rendered according to the poses of the first scanlines, via overlaying.
The red and blue color regions indicate high differences.
Compared to the original RS images, one can see that the rectified images from our RS-aware warping are closer to the ground truth GS images,
except in the few regions near the image edges where the optical flow computation may not be reliable.
In Fig.~\ref{fig:ptcloudsyn2}, we show the point clouds reconstructed by the GS model, our
RS model, and the ground truth respectively.
As highlighted by the boxes, the point clouds returned by the GS model are distorted compared to the ground truth. In comparison, our RS model successfully rectifies these artifacts to
obtain visually more appealing results.

\subsection{Real data}
\vspace{-2mm}
\begin{figure}
 \setlength{\abovecaptionskip}{-0.2cm}
 \setlength{\belowcaptionskip}{-0.6cm}
\begin{center}
   \includegraphics[width=1\linewidth, trim = 40mm 126mm 98mm 15mm, clip]{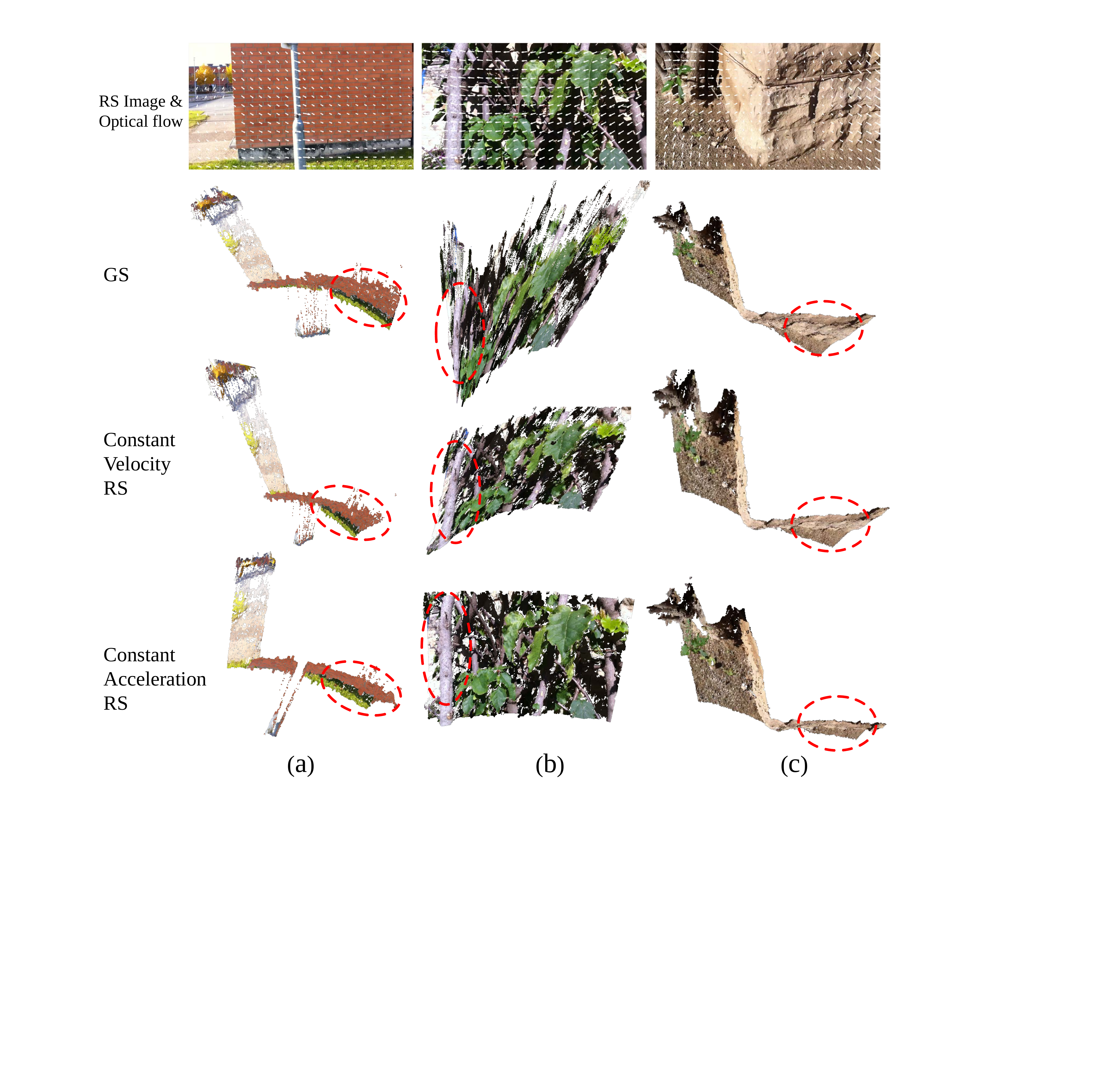}
\end{center}
\vspace{-1.5mm}
   \caption{SfM results on real image data. Top row: original RS images. Bottom 3 rows: reconstructed 3D point clouds by the GS model
   and our RS models with constant velocity and acceleration.}
\label{fig:ptcloudreal}
\end{figure}
In this section, we show the results of applying the proposed RS algorithm to images collected by real RS cameras.
First, we show the results on pairs of consecutive images from the public RS images dataset released by \cite{hedborg2012rolling}. The sequence was collected by an
Iphone 4 camera at $1280\times720$ resolution with 96\% readout time ratio. Despite having a GS camera
that is rigidly mounted near the Iphone for ground truth comparison over long trajectories as shown in \cite{hedborg2012rolling},
the accuracy is insufficient for the images from the GS camera to be used as ground truth for two-frame differential relative pose estimation.
Instead, we rely on the visual quality of the reconstructed point clouds to evaluate our algorithms.
We show the point clouds of three different scenes by the GS and our RS models---both constant velocity and acceleration in Fig.~\ref{fig:ptcloudreal}. More results are shown in \textit{supplementary material}.
 As highlighted by the red ellipses in Fig.~\ref{fig:ptcloudreal}(a), we can see from the top-down view that the wall is significantly skewed under the GS model.
 This distortion is corrected to a certain extent and almost completely removed by our constant velocity and acceleration RS models respectively.
 Similar performance of our RS models can also be observed in  the examples shown in  Fig.~\ref{fig:ptcloudreal}(b) and Fig.~\ref{fig:ptcloudreal}(c) from front and top-down view respectively.

\begin{figure*}
 \setlength{\abovecaptionskip}{-0.25cm}
 \setlength{\belowcaptionskip}{-0.45cm}
\begin{center}
   \includegraphics[width=0.92\linewidth, trim = 30mm 60mm 80mm 34mm, clip]{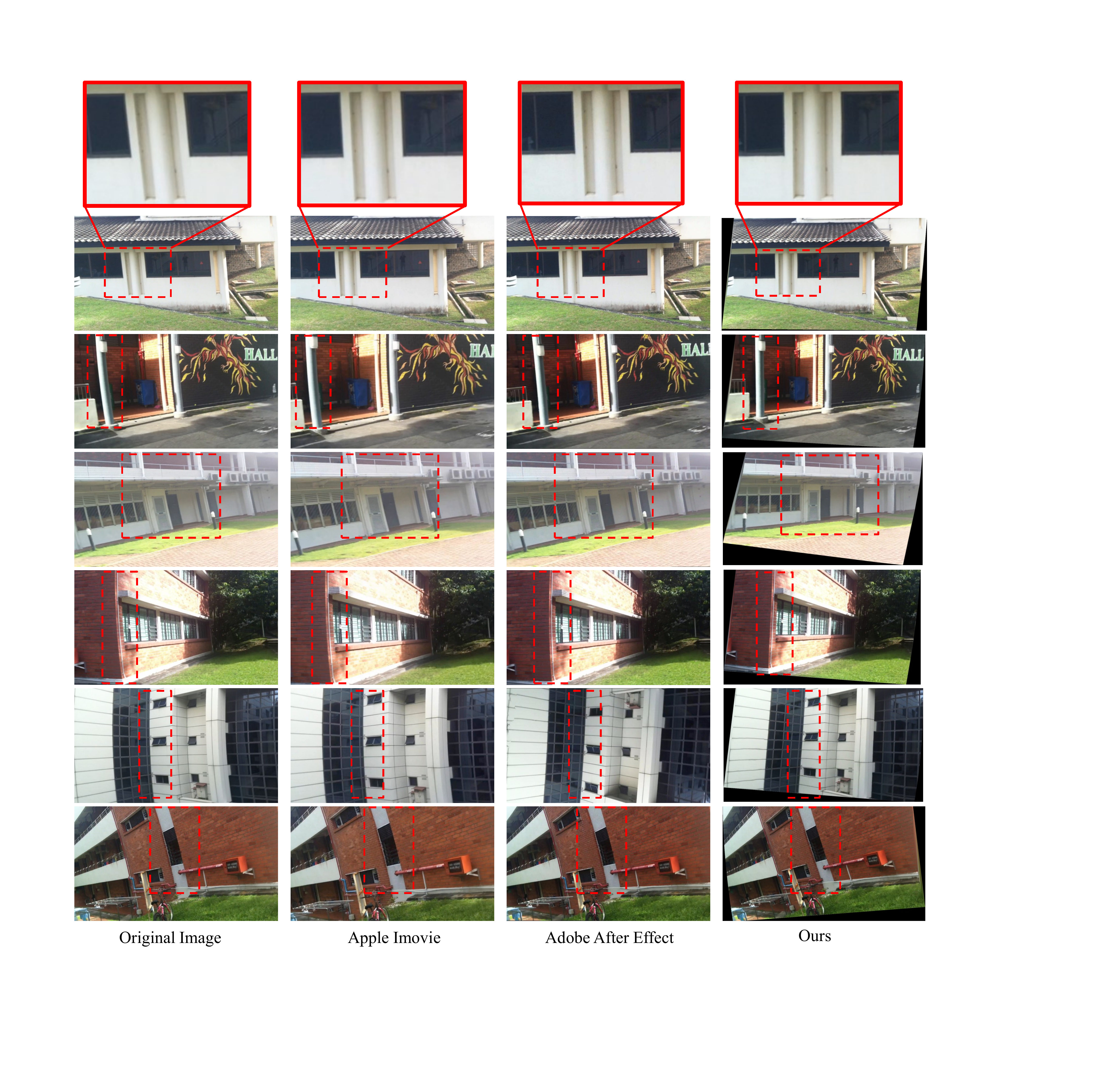}
\end{center}
   \caption{Comparison of image rectification results on real image data with noticeable RS distortion. The red boxes highlight the superior performance of our proposed method.}
\label{fig:undistcomparison}
\end{figure*}

The RS effect from the above mentioned dataset is significant enough to introduce bias in SfM algorithm, but
it is not strong enough to generate noticeable image distortions. To demonstrate our image rectification algorithm, we
collected a few image sequences with an Iphone 4 camera under larger motions that lead to obvious RS distortions on the images.
We compare the results of our proposed method with those of Rolling Shutter Repair in two image/video editing software products---Adobe After Effect and Apple Imovie on pairs of the collected images, as
shown in Fig.~\ref{fig:undistcomparison}.
We feed the image sequences along with camera parameters into the software products. We tried different advanced settings provided by After Effect to get the best result for each scene.
Since we observe that both our RS models with constant velocity or acceleration give similar results, we only report the rectified images using the accelerated motion model.
It can be seen that our method works consistently better than the two commercial software products in removing the RS artifacts such as skew and wobble in the images (highlighted by the red boxes).
For example, the slanted window on the original RS image shown on the top row of Fig.~\ref{fig:undistcomparison} becomes most close to vertical in our result.

\vspace{-0mm}

\section{Conclusion}

In this paper, we proposed two tractable algorithms to
\newpage
 \noindent correct the inaccuracies in differential SfM caused by the RS effect in
images collected from a RS camera moving under constant velocity and acceleration respectively.
In addition, we proposed the use of a RS-aware warping for image rectification that removes the RS distortion on images.
Quantitative and qualitative experimental results on both synthetic and real RS images demonstrated the effectiveness of our algorithm.\\
 \noindent \textbf{Acknowledgements}. This work was partially supported by the Singapore PSF grant 1521200082 and Singapore MOE Tier 1 grant R-252-000-636-133.

{\small
\bibliographystyle{ieee}
\bibliography{egbib}
}

\end{document}